\documentclass{article}

\usepackage{arxiv}

\usepackage{graphics} % for pdf, bitmapped graphics files
\usepackage{amsmath} % assumes amsmath package installed
\usepackage{mathtools} % left align
\usepackage[utf8]{inputenc} % allows for new lines
\usepackage[english]{babel}
\usepackage[numbers, sort, comma, square]{natbib}

\usepackage{helvet}         % selects Helvetica as sans-serif font
\usepackage{courier}        % selects Courier as typewriter font
\usepackage{type1cm}        % activate if the above 3 fonts are
\usepackage{makeidx}         % allows index generation
\usepackage{graphicx}        % standard LaTeX graphics tool
\usepackage{multicol}        % used for the two-column index
\usepackage[bottom]{footmisc}% places footnotes at page bottom
\usepackage{subfig}  %if using subfloat, not compatible with subcaption package.

\usepackage{tikz}
\usetikzlibrary{positioning}
\usetikzlibrary{shapes,snakes}
\usepackage{bm} % allows use of \bm command in math mode
\usepackage{float} % allow using of [H] tag on "figure"s
\usepackage{color} % color text package

\usepackage{caption}
\usepackage{amsmath,amssymb} % allows for bmatrix etc.
\usepackage{graphicx}

\usepackage{empheq} % Boxed equations http://tex.stackexchange.com/a/14298
\usepackage{xcolor} % Boxed equations http://tex.stackexchange.com/a/14298

\newcommand\scalemath[2]{\scalebox{#1}{\mbox{\ensuremath{\displaystyle #2}}}}

\title{Visual-Inertial Navigation: A Concise Review}
\author{Guoquan (Paul) Huang \vspace{5pt}\\% <-this % stops a space
Robot Perception and Navigation Group\\
University of Delaware, Newark, DE 19716\\
Email: {\tt ghuang@udel.edu}
}

\begin{document}

\maketitle

%----------------------------------------------------------------------------------------------
\begin{abstract}
% \lipsum[1]
As inertial and visual sensors are becoming ubiquitous,  
visual-inertial navigation systems (VINS) have  prevailed in a wide range of applications 
from mobile augmented reality to aerial navigation to autonomous driving,
in part because of the complementary sensing capabilities and the decreasing costs and size of the sensors. 
In this paper, we survey thoroughly the research efforts taken in this field
and strive to provide a concise but complete review of the related work --
which is unfortunately missing in the literature while being greatly demanded by researchers and engineers --
in the hope to accelerate the VINS research and beyond in our society as a whole.
\end{abstract}

%-----------------------------------------------------------------------------------------------------
\section{Introduction} \label{sec:intro}

Over the years, inertial navigation systems (INS)~\cite{Chatfield1997,Titterton2005} have been widely used for
estimating the 6DOF poses (positions and orientations) of sensing platforms (e.g., autonomous vehicles), in particular, in GPS-denied environments
such as underwater, indoor, in the urban canyon, and on other planets.
Most INS rely on a 6-axis inertial measurement unit (IMU)
that measures the local linear acceleration and angular velocity  of the platform to which it is rigidly connected.
With the recent advancements of hardware design and manufacturing, 
low-cost light-weight micro-electro-mechanical (MEMS) IMUs have become ubiquitous~\cite{Maenaka2008ICNSS,Barbour2010TR,Ahmad2013IJSPS},
which  enables high-accuracy localization for, among others, mobile devices~\cite{Wu2015RSS} and micro aerial vehicles (MAVs)~\cite{Shen2014ISER,Shen2014thesis,Ling2016ICRA,Do2018IJRR,Delmerico2018ICRA},
holding huge implications in a wide range of emerging applications from mobile augmented reality (AR)~\cite{arcore,arkit} and virtual reality (VR)~\cite{oculus} to autonomous driving~\cite{Wu2017ICRA,Bresson2017TIV}.
Unfortunately, simple integration of high-rate IMU measurements that are corrupted by noise and bias,
often results in pose estimates unreliable for long-term navigation.
Although a high-end tactical-grade IMU exists, it remains prohibitively expensive for widespread deployments.
On the other hand,
a camera  that is small, light-weight, and energy-efficient,
provides rich information about the environment and  serves as an idea aiding source for INS,
yielding visual-inertial navigation systems (VINS).

While this problem is challenging because of the lack of global information to reduce the motion drift accumulated over time  
(which is even exacerbated if low-cost, low-quality sensors are used),
VINS have attracted significant attentions over the last decade.
To date, many VINS algorithms are available for both visual-inertial SLAM~\cite{Kim2007RAS} 
and visual-inertial odometry (VIO)~\cite{Mourikis2007ICRA,Li2012ICRA}, 
such as the extended Kalman filter (EKF)~\cite{Kim2007RAS,Bryson2008TAES,Mourikis2007ICRA,Mourikis2009TRO,Huai2018IROS}, 
the unscented Kalman filter (UKF)~\cite{Ebcin2007TR,Loianno2016ICRA,Brossard2018FUSION}, 
the batch or incremental smoother~\cite{Strelow2004,Indelman2013RAS},
and (window) optimization-based approaches~\cite{Leutenegger2014IJRR,Forster2017TRO,Eckenhoff2019IJRR,Qin2018TRO}.
Among these, the EKF-based methods remain popular because of its efficiency.
For example, as a state-of-the-art solution of VINS on mobile devices,  Project Tango~\cite{Tango} (or ARCore~\cite{arcore}) appears to use an EKF
to fuse the visual and inertial measurements for motion tracking.
%by processing them in a single estimation thread.
%%
Nevertheless,  recent advances of preintegration have also allowed for efficient inclusion of high-rate IMU measurements in graph optimization-based  formulations~\citep{Lupton2012TRO,Forster2015RSS,Forster2017TRO,Eckenhoff2016WAFR,Eckenhoff2019IJRR}.

As evident, VINS technologies are emerging, largely due to the demanding mobile perception/navigation applications,
which has given rise to a rich body of literature in this area.
However, to the best of out knowledge, there is {\em no} contemporary literature review of VINS, 
although there are recent surveys broadly about SLAM~\citep{Cadena2016TRO,Bresson2017TIV} while not specializing on VINS.
This has made difficult for researchers and engineers in both academia and industry, 
to effectively find and understand the most important related work to their interests,
which we have experienced over the years when we are working on this problem.
For this reason, we are striving to bridge this gap by:
(i)~offering a concise  (due to space limitation) but complete review on VINS while focusing on the key aspects of state estimation,
(ii)~providing our understandings about the most important related work,
and (iii)~opening discussions about the challenges remaining to tackle.
This is driven by the hope to (at least) help researchers/engineers  track and understand the state-of-the-art VINS algorithms/systems,  
more efficiently and effectively,
thus accelerating the VINS research and development  in our society as a whole.
%

%-----------------------------------------------------------------------------------------------------
\section{Visual-Inertial Navigation}  \label{sec:vins}

In this section,  we provide some basic background of canonical VINS,
by describing the IMU propagation and camera measurement models 
within the  EKF framework.

\subsection{IMU Kinematic Model} \label{sec:imu-model}

The EKF uses the IMU (gyroscope and accelerometer) measurements for state propagation,
and the state vector consists of the IMU states $\mathbf x_I $ and the feature position $^G\mathbf p_f$:
% \footnote{Throughout this paper 
% the subscript $\ell |j$ refers to the estimate of a
% quantity at time-step $\ell$, after all measurements up to time-step $j$ have been processed. $\hat
% x$ is used to denote the estimate of a random variable $x$, while $\tilde x = x-\hat x$ is  the error in this estimate. 
% $\mathbf I_n$ and $\mathbf 0_n$ are the $n \times n$ identity and zero matrices, respectively.
% Finally, the left superscript denotes the frame of reference with respect to which the vector is expressed.}
%
\begin{align} \label{eq:imu-state}
\mathbf x 
&= \begin{bmatrix} \mathbf x_I^T & ^G\mathbf p_f^T \end{bmatrix}^T  \notag\\
&= \begin{bmatrix} ^I_G\bar{\mathbf q}^T & \mathbf b_g^T & ^G\mathbf v^T & \mathbf b_a^T & ^G\mathbf p^T &  ^G\mathbf p_f^T  \end{bmatrix}^T
\end{align}
where $^I_G\bar{\mathbf q}$ is the unit quaternion that represents the rotation from the global frame of reference $\{G\}$ to the IMU frame $\{I\}$ 
(i.e., different parametrization of the rotation matrix $\mathbf C(^I_G\bar{\mathbf q}) =: {^I_G\mathbf C}$);
$^G\mathbf p$ and $^G\mathbf v$ are the IMU position and velocity in the global frame;
and $\mathbf b_g$ and $\mathbf b_a$ denote the gyroscope and accelerometer biases, respectively.

By noting that the feature is static (with trivial dynamics), 
as well as using the IMU motion dynamics~\cite{Trawny2005_Q_TR},
the continuous-time dynamics of the state~\eqref{eq:imu-state} is given by:
\begin{align}
^I_G\dot{\bar{\mathbf q}}(t) &=\! \frac{1}{2} \bm\Omega\left(^I\bm\omega(t)\right) {^I_G\bar{\mathbf q}}(t),~ 
^G\dot{\mathbf p}(t) =\! {^G\mathbf v(t)},~
^G\dot{\mathbf v}(t) =\! {^G\mathbf a}(t) \notag\\
\dot {\mathbf b}_g(t) &= \mathbf n_{wg}(t) ,~
\dot {\mathbf b}_a(t) = \mathbf n_{wa}(t),~
^G\dot{\mathbf p}_f(t) = \mathbf 0_{3\times 1} 
\label{eq:imu-motion}
\end{align}
where $^I\bm\omega = \begin{bmatrix} \omega_1 & \omega_2 & \omega_3 \end{bmatrix}^T$ is the rotational velocity of the IMU, 
expressed in $\{I\}$, $^G\mathbf a$ is the IMU acceleration in $\{G\}$, 
$\mathbf n_{wg}$ and $\mathbf n_{wa}$ are the white Gaussian noise processes that drive the IMU biases,
%and $\bm\Omega(\bm\omega)$ is defined by:
%\begin{align*}
and $\bm\Omega(\bm\omega) = \begin{bmatrix}
-\lfloor \bm\omega \times \rfloor & \bm\omega \\
-\bm\omega^T & 0 \end{bmatrix}$, where $\lfloor \bm\omega \times \rfloor$ is the skew-symmetric matrix.
%
%\lfloor \bm\omega \times \rfloor = \begin{bmatrix}
%0 & -\omega_3 & \omega_2 \\
%\omega_3 & 0 & -\omega_1 \\
%-\omega_2 & \omega_1 & 0
%\end{bmatrix}
%\end{align*}
%

A typical IMU provides gyroscope and accelerometer measurements, $\bm\omega_m$ and $\mathbf a_m$, 
both of which are expressed in the IMU local frame $\{I\}$ and given by:
\begin{align}
\bm\omega_m(t) &= {^I\bm\omega}(t) + \mathbf b_g(t) + \mathbf n_g(t) \\
\mathbf a_m(t) &= \mathbf C(^I_G\bar{\mathbf q}(t))  \left( {^G\mathbf a}(t) - {^G\mathbf g}  \right) + \mathbf b_a(t) + \mathbf n_a(t)
\end{align}
where ${^G\mathbf g}$ is the gravitational acceleration expressed in $\{G\}$, 
and $\mathbf n_g$ and $\mathbf n_a$ are zero-mean, white Gaussian noise.

Linearization of~\eqref{eq:imu-motion} at the current state estimate yields 
the continuous-time state-estimate propagation model~\cite{Mourikis2009TRO}:
\begin{align}
^I_G\dot{\hat{\bar{\mathbf q}}}(t) &=\! \frac{1}{2} \bm\Omega\left(^I\hat{\bm\omega}(t)\right) {^I_G\hat{\bar{\mathbf q}}}(t),~ 
^G\dot{\hat{\mathbf p}}(t) =\! {^G\hat{\mathbf v}(t)} ,~
^G\dot{\hat{\mathbf v}}(t) =\! {^G\hat{\mathbf a}(t)} \notag\\
\dot{\hat{\mathbf b}}_g(t) &= \mathbf 0_{3\times 1} ,~
\dot {\hat{\mathbf b}}_a(t) = \mathbf 0_{3\times 1} ,~
^G\dot{\hat{\mathbf p}}_f(t) = \mathbf 0_{3\times 1} 
\label{eq:imu-prop}
\end{align}
where $\hat{\mathbf a} = \mathbf a_m - \hat{\mathbf b}_a$ and $\hat{\bm\omega} = \bm\omega_m - \hat{\mathbf b}_g$.
The error state of dimension $18\times 1$ is hence defined as follows [see~\eqref{eq:imu-state}]:
\begin{align} \label{eq:imu-err-state}
\scalemath{.9}{
\widetilde{\mathbf x}(t) = 
\begin{bmatrix}
^I\widetilde{\bm\theta}^T(t)  \!&\! \mathbf{\widetilde b}_g^T(t) \!&\!  ^G\mathbf{\widetilde v}^T(t) \!&\!  \mathbf{\widetilde b}_a^T(t) 
\!&\! ^G\mathbf{\widetilde p}^T(t) \!&\!  ^G\mathbf{\widetilde p}_f^T(t) 
\end{bmatrix}^T
}
\end{align}
where we have employed the multiplicative error model for a quaternion~\cite{Trawny2005_Q_TR}.
That is, the error between the quaternion $\bar{\mathbf q}$ and its estimate $\hat{\bar{\mathbf q}}$ is the $3\times 1$ angle-error vector,
$^I\widetilde{\bm\theta}$, implicitly defined by the error quaternion: 
$\delta\bar{\mathbf q} = \bar{\mathbf q} \otimes \hat{\bar{\mathbf q}} \simeq \begin{bmatrix}
\frac{1}{2} {^I\widetilde{\bm\theta}} \\ 1 \end{bmatrix}$,
where $\delta\bar{\mathbf q}$ describes the small rotation that causes the true and estimated attitude to coincide.
The advantage of this parametrization permits a minimal representation, 
$3\times 3$ covariance matrix $\mathbb{E}\left[ {^I\widetilde{\bm\theta}} ~ {^I\widetilde{\bm\theta}}^T\right]$,
for the attitude uncertainty.
%
%It is important to note that the orientation error, $^I\widetilde{\bm\theta}$, 
%satisfies the following rotation-matrix relation~\cite{Trawny2005_Q_TR}:
%\begin{align} \label{eq:local-rot}
%\mathbf C(^I_G\bar{\mathbf q}) \simeq \left( \mathbf I_3 - \lfloor ^I\widetilde{\bm\theta} \times\rfloor \right) 
%\mathbf C(^I_G\hat{\bar{\mathbf q}})
%\end{align}

Now the continuous-time error-state propagation is:
\begin{align}
\dot{\widetilde{\mathbf x}}(t) = 
\mathbf F_c(t) \widetilde{\mathbf x}(t) + \mathbf G_c(t) \mathbf n(t)
\end{align}
where 
$\mathbf n = \begin{bmatrix} \mathbf n_g^T & \mathbf n_{wg}^T & \mathbf n_a^T & \mathbf n_{wa}^T \end{bmatrix}^T$
is the system noise, 
$\mathbf F_c$ is the continuous-time error-state transition matrix, 
and $\mathbf G_c$ is the input noise matrix, which are given by (see~\cite{Trawny2005_Q_TR}):
\begin{align}
\mathbf F_c &= 
\scalemath{.88}{
\begin{bmatrix}
-\lfloor \hat{\bm\omega} \times \rfloor & -\mathbf I_3 & \mathbf 0_3 & \mathbf 0_3 & \mathbf 0_3 & \mathbf 0_3\\
\mathbf 0_3 & \mathbf 0_3 & \mathbf 0_3 & \mathbf 0_3 & \mathbf 0_3 & \mathbf 0_3 \\
-\mathbf C^T(^I_G \hat{\bar{\mathbf q}}) \lfloor \hat{\mathbf a}\times\rfloor &  \mathbf 0_3  & \mathbf 0_3 & -\mathbf C^T(^I_G \hat{\bar{\mathbf q}})  & \mathbf 0_3 & \mathbf 0_3 \\
\mathbf 0_3 & \mathbf 0_3 & \mathbf 0_3 & \mathbf 0_3 & \mathbf 0_3 & \mathbf 0_3 \\
\mathbf 0_3 & \mathbf 0_3 & \mathbf I_3 & \mathbf 0_3 & \mathbf 0_3 & \mathbf 0_3 \\
\mathbf 0_3 & \mathbf 0_3 & \mathbf 0_3 & \mathbf 0_3 & \mathbf 0_3 & \mathbf 0_3 
\end{bmatrix}
} \label{eq:Fc}  \\
\mathbf G_c & = 
\scalemath{.95}{
\begin{bmatrix}
-\mathbf I_3 & \mathbf 0_3 & \mathbf 0_3  & \mathbf 0_3  \\
-\mathbf 0_3 & \mathbf I_3 & \mathbf 0_3  & \mathbf 0_3  \\
-\mathbf I_3 & \mathbf 0_3 &  -\mathbf C^T(^I_G \hat{\bar{\mathbf q}})  & \mathbf 0_3  \\
-\mathbf I_3 & \mathbf 0_3 & \mathbf 0_3  & \mathbf I_3  \\
-\mathbf I_3 & \mathbf 0_3 & \mathbf 0_3  & \mathbf 0_3  \\
\mathbf 0_3 & \mathbf 0_3 & \mathbf 0_3 & \mathbf 0_3 
\end{bmatrix} 
} \label{eq:Gc}
\end{align}
The system noise is modelled as zero-mean white Gaussian process with autocorrelation 
$\mathbb E \left[ \mathbf n(t) \mathbf n(\tau)^T \right] = \mathbf Q_c \delta(t-\tau)$,
which depends on the IMU noise characteristics. %and is calibrated off-line~\cite{Trawny2005_Q_TR}.

%%%%%%%%%%%%
We have described the continuous-time propagation model using IMU measurements. 
However, in any practical EKF implementation, 
the discrete-time state-transition matrix, $\bm\Phi_k := \bm\Phi(t_{k+1},t_{k})$, 
is required in order to propagate the error covariance from time $t_{k}$ to $t_{k+1}$.
Typically it is found by solving the following matrix differential equation:
\begin{align} \label{eq:mat-diff}
\dot{\bm\Phi} (t_{k+1},t_k) = \mathbf F_c(t_{k+1}) \bm\Phi (t_{k+1},t_k)
\end{align}
with the initial condition $\bm\Phi(t_{k},t_k) = \mathbf I_{18}$.
%
%Its solution has the following structure:
%\begin{align}\label{eq:imu-Phi}
%\scalemath{.925}{
%\bm\Phi_k := \bm\Phi(t_{k+1},t_k) = 
%\begin{bmatrix}
%\bm\Phi_{k,{11}} & \bm\Phi_{k,{12}} & \mathbf 0_3  & \mathbf 0_3  & \mathbf 0_3 & \mathbf 0_3\\
%\mathbf 0_3 & \mathbf I_3  & \mathbf 0_3 & \mathbf 0_3 & \mathbf 0_3 & \mathbf 0_3 \\
%\bm\Phi_{k,{31}}  & \bm\Phi_{k,{32}}  & \mathbf I_3  & \bm\Phi_{k,{34}} & \mathbf 0_3  & \mathbf 0_3 \\
%\mathbf 0_3 & \mathbf 0_3  & \mathbf 0_3 & \mathbf I_3 & \mathbf 0_3 & \mathbf 0_3\\
%\bm\Phi_{k,{51}} & \bm\Phi_{k,{52}} & \delta t_k \mathbf I_3  & \bm\Phi_{k,{54}} & \mathbf I_3 & \mathbf 0_3 \\
%\mathbf 0_3 & \mathbf 0_3  & \mathbf 0_3 & \mathbf 0_3 & \mathbf 0_3 & \mathbf I_3 
%\end{bmatrix} }
%\end{align}
%where $\delta t_k = t_{k+1} -t_k$.
%This matrix~\eqref{eq:imu-Phi} 
This can be solved either numerically~\cite{Trawny2005_Q_TR,Mourikis2009TRO} 
or analytically~\cite{Hesch2013WAFR,Hesch2013TRO,Li2012ICRA,Li2013IJRR}.
Once it is computed, 
the EKF propagates the covariance as~\cite{Maybeck1979}:
\begin{align} \label{eq:cov-prop}
\mathbf P_{k+1|k} = \bm\Phi_k \mathbf P_{k|k} \bm\Phi_k^T + \mathbf Q_{d,k}
\end{align}
where $\mathbf Q_{d,k}$ is the discrete-time system noise covariance matrix computed as follows:
\begin{align*}
\mathbf Q_{d,k} = \int_{t_{k}}^{t_{k+1}} \bm\Phi(t_{k+1},\tau) \mathbf G_c(\tau) \mathbf Q_c \mathbf G_c^T(\tau)  \bm\Phi^T(t_{k+1},\tau) d\tau
\end{align*}
%

%%%%%%%%%%%%%%%%%%%%%%%%%%%%%%%%%%%%%%%%%
\subsection{Camera Measurement Model} \label{sec:meas-model}

The camera observes visual corner features,
which are used to concurrently estimate the ego-motion of the sensing platform.
Assuming a calibrated perspective camera, the measurement of the feature at time-step $k$ is 
the perspective projection of the 3D point, 
$^{C_k}\mathbf p_{f}$, expressed in the current camera frame $\{C_k\}$, 
onto the image plane, i.e.,
\begin{align} \label{eq:meas-model}
\mathbf z_{k} &= \frac{1}{z_k} \begin{bmatrix} x_k \\ y_k \end{bmatrix} + \mathbf n_{f_k} \\
\begin{bmatrix} x_k \\ y_k \\ z_k  \end{bmatrix} &= 
\scalemath{.95}{
{^{C_k}\mathbf p_{f}}  
= \mathbf C(^C_I \bar{\mathbf q}) \mathbf C(^I_G \bar{\mathbf q}_k) \left({^G\mathbf p_{f}} - {^G\mathbf p_{k}} \right) + {^C\mathbf p_I}
}
\label{eq:meas-eq}
\end{align}
where $\mathbf n_{f_k}$ is the zero-mean, white Gaussian measurement noise with covariance $\mathbf R_{k}$.
In~\eqref{eq:meas-eq}, $\{^C_I \bar{\mathbf q},  {^C\mathbf p_I} \}$ 
is the rotation and translation between the camera and the IMU.
This transformation can be obtained, for example, 
by performing camera-IMU extrinsic calibration {\em offline}~\cite{Mirzaei2008TRO}.
%%%
%%%
However, in practice when the perfect calibration is unavailable, 
it is beneficial to VINS consistency to 
include these calibration parameters in the state vector 
and concurrently estimate them along with the IMU/camera poses~\cite{Li2013IJRR}. 
%For this reason, we perform {\em online} camera-IMU calibration in the proposed STOC-VINS (see Section~\ref{sec:stoc-vins}).
%%%%%

For the use of EKF, linearization of~\eqref{eq:meas-model}  yields 
the following measurement residual [see~\eqref{eq:imu-err-state}]:
\begin{align} \label{eq:residual}
\scalemath{.9}{
\widetilde{\mathbf z}_{k} 
= \mathbf H_{k} \widetilde{\mathbf x}_{{k|k-1}} \!+ \mathbf n_{f_k} 
= \mathbf H_{\mathbf I_{k}} \widetilde{\mathbf x}_{I_{k|k-1}} \!+ \mathbf H_{\mathbf f_{k}} {^G\widetilde{\mathbf p}_{f_{k|k-1}}} \!+ \mathbf n_{f_k}
}
\end{align}
where the measurement Jacobian $\mathbf H_{k}$ is computed as:
\begin{align} \label{eq:meas-jac}
&\scalemath{.9}{
\mathbf H_{k} = \begin{bmatrix} \mathbf H_{\mathbf I_{k}} &   \mathbf H_{\mathbf f_{k}}   \end{bmatrix} 
} \\
&\scalemath{.9}{
~~~~= \mathbf {H_{proj}} \mathbf C(^C_I \bar{\mathbf q}) \begin{bmatrix} \mathbf H_{\bm\theta_k} & \mathbf 0_{3\times 9} & \mathbf H_{\mathbf p_k}  &&  \mathbf C(^I_G \hat{\bar{\mathbf q}}_k) \end{bmatrix} 
} \notag\\
%\mathbf H_{\mathbf I_{k}} &= \mathbf {H_{proj}} \mathbf C(^C_I \bar{\mathbf q}) \begin{bmatrix} \mathbf H_{\bm\theta_k} & \mathbf 0_{3\times 9} & \mathbf H_{\mathbf p_k} \end{bmatrix} \\
%\mathbf H_{\mathbf f_{k}} &=  \mathbf {H_{proj}} \mathbf C(^C_I \bar{\mathbf q})  \mathbf C(^I_G \bar{\mathbf q}_k) \\
%
&\scalemath{.9}{
\mathbf {H_{proj}} = \frac{1}{\hat z_k^2} \begin{bmatrix} \hat z_k & 0 & -\hat x_k \\ 0 & \hat z_k & -\hat y_k \end{bmatrix}  
}\\
&\scalemath{.9}{
\mathbf H_{\bm\theta_k} = \lfloor  \mathbf C(^I_G \hat{\bar{\mathbf q}}_k) \left({^G\hat{\mathbf p}_{f}} - {^G\hat{\mathbf p}_{k}} \right) \times  \rfloor  ~,~
\mathbf H_{\mathbf p_k} = -\mathbf C(^I_G \hat{\bar{\mathbf q}}_k) 
}
\label{eq:meas-jac2}
\end{align}
%
%%%
Once the measurement Jacobian and residual are computed, 
we can apply the standard EKF update equations to update the state estimates and error covariance~\cite{Maybeck1979}.

%-----------------------------------------------------------------------------------------------------
\section{State Estimation} \label{sec:estimation}

It is clear from the preceding section that
at the core of visual-inertial navigation systems (VINS) is a state estimation algorithm [see \eqref{eq:imu-motion} and \eqref{eq:meas-model}],
aiming to optimally fuse IMU measurements and camera images to provide motion tracking of the sensor platform.
In this section, we review the VINS literature by focusing on the estimation engine.

\subsection{Filtering-based vs. Optimization-based Estimation}

\citet{Mourikis2007ICRA} developed one of the earliest successful VINS algorithms, known as the multi-state constraint Kalman filter (MSCKF),
which later was applied to the application of spacecraft  descent and landing~\cite{Mourikis2009TRO} and fast UAV autonomous flight~\cite{Sun2018RAL}.
This approach uses the quaternion-based inertial dynamics~\cite{Trawny2005_Q_TR} for state propagation tightly coupled with an efficient EKF update.
Specifically, rather than adding features detected and tracked over the camera images to the state vector, their visual bearing measurements are projected onto the null space of the feature Jacobian matrix (i.e., linear marginalization~\cite{Yang2017IROS}), 
thereby retaining motion constraints that only relate to the stochastically cloned camera poses in the state vector~\cite{Roumeliotis2002ICRAa}.
While reducing the computational cost by removing the need to co-estimate potentially hundreds  and thousands of point features, 
this operation prevents the relinearization of the features' nonlinear measurements at later times, 
yielding approximations deteriorating its performance. 
The standard MSCKF~\cite{Mourikis2007ICRA} recently has been extended and improved along different directions.
In particular, 
by exploiting the observability-based methodology proposed in our prior work~\cite{Huang2008ICRA,Huang2008ISER,Huang2010IJRR,Huang2012thesis},
different observability-constrained (OC)-MSCKF algorithms have been developed to improve the filter consistency 
by enforcing the correct observability properties of the linearized VINS~\cite{Kottas2012ISER,Hesch2013WAFR,Hesch2013TRO,Hesch2014IJRR,Li2012ICRA,Li2013IJRR,Huang2014ICRA}.
A square-root inverse version of the MSCKF, i.e., the square-root inverse sliding window filter (SR-ISWF)~\cite{Wu2015RSS,Paul2017ICRA} was introduced  to improve the computational efficiency and numerical stability to enable VINS running on mobile devices with limited resources while not sacrificing estimation accuracy.
We have  introduced the optimal state constraint (OSC)-EKF~\cite{Huang2015ISRR,Maley2017TR} that  first optimally extracts all the information contained in the visual measurements about the relative camera poses in a sliding window and then uses these inferred relative-pose measurements in the EKF update.
The  (right) invariant Kalman filter~\cite{Barrau2018ARCRAS} was recently employed to improve filter consistency~\cite{Wu2017IROS,Zhang2017RAL,Brossard2018FUSION,Heo2018ISJ,Brossard2017IROS}, 
as well as the (iterated) EKF that was also used for VINS in robocentric formulations~\cite{Bloesch2015IROS,Bloesch2017IJRR,Huai2018IROS,Huai2019IJRR}.
On the other hand, 
in the EKF framework, 
different geometric features besides points have also been exploited to improve VINS performance, for example, 
line features used in~\cite{Yu2015IROS,Kottas2013ICRA,Heo2018Sensors,Zheng2018ICRA,He2018Sensors}
and plane features in~\cite{Guo2013iros,Hsiao2018ICRA,Geneva2018IROS,Yang2019ICRAb}.
In addition, 
the MSCKF-based VINS was also extended to use rolling-shutter cameras with inaccurate time synchronization~\cite{Guo2014RSS,Yu2015IROS}, RGBD cameras~\cite{Guo2013iros,Guo2013ICRA}, multiple cameras~\cite{Paul2017ICRA,Paul2018CVPR,Eckenhoff2019ICRAa} and multiple IMUs~\cite{Eckenhoff2019ICRAb}.
While the filtering-based VINS have shown to exhibit high-accuracy state estimation, they theoretically suffer from a limitation; 
that is, nonlinear measurements~\eqref{eq:meas-model} must have a \textit{one-time} linearization before processing, 
possibly introducing large linearization errors into the estimator and degrading performance.

%%%%%%%
Batch optimization methods, by contrast, solve a nonlinear least-squares (bundle adjustment or BA~\cite{Triggs2000VA}) problem over a set of measurements, allowing for the reduction of error through relinearization~\citep{Dellaert2006IJRR,Kuemmerle2011ICRA} but with high computational cost.
\citet{Indelman2013RAS} employed the  factor graph to represent the VINS problem and then solved it incrementally in analogy to iSAM~\cite{Kaess2008TRO,Kaess2012IJRR}.
To achieve constant processing time when applied to VINS, typically a bounded-size sliding window of recent states are only considered as active optimization variables 
while marginalizing out past states and measurements~\cite{Leutenegger2014IJRR,Mur-Artal2017RAL,Yang2017TASE,Shen2015ICRA,Qin2018TRO}.
%Due to high computational demands of iterative solving of nonlinear systems, few graph-based can achieve real-time performance on resource-constrained platforms, such as mobile phones.
%
In particular, \citet{Leutenegger2014IJRR} introduced a keyframe-based optimization approach (i.e., OKVIS), whereby a set of non-sequential past camera poses and a series of recent inertial states, connected with inertial measurements, was used in nonlinear optimization for accurate trajectory estimation.
%These inertial factors took the form of a state prediction: every time that the linearization point for the starting inertial state changed, it is  required to reintegrate the IMU dynamics.
%This presents inefficiencies in the inertial processing, and makes incorporating a large number of inertial factors practically infeasible.
%
\citet{Qin2018TRO} recently presented an optimization-based monocular VINS that can incorporate loop closures in a non-real time thread,
while our recent VINS~\cite{Geneva2019ICRA} is able to efficiently utilize loop closures in a single thread with linear complexity.

\subsection{Tightly-coupled vs. Loosely-coupled Sensor Fusion}

There are different schemes  for VINS to fuse the visual and inertial measurements 
which can be broadly categorized into the {loosely-coupled} and the {tightly-coupled}. 
Specifically, 
the loosely-coupled fusion,
in either filtering or optimization-based estimation,
processes the visual and inertial measurements separately to infer their own motion constraints 
and then fuse these constraints (e.g., \cite{Klei2007ISMAR,Weiss2011ICRA,Weiss2012ICRAb,Lynen2013IROS,Kneip2011BMVC,Indelman2013RAS}). 
Although this method is computationally efficient, the decoupling of visual and inertial constraints results in information loss. 
By contrast, the tightly-coupled approaches directly fuse the visual and inertial measurements within a single process, 
thus  achieving  higher accuracy (e.g., \cite{Mourikis2007ICRA,Li2013IJRR,Forster2015RSS,Shen2015ICRA,Leutenegger2014IJRR,Eckenhoff2019RAL}). 

\subsection{VIO vs. SLAM}

By jointly estimating the location of the sensor platform and the features in the surrounding environment, 
SLAM estimators are able to easily incorporate loop closure constraints, thus enabling bounded localization errors,
which has attracted significant research efforts in the past three decades~\cite{Durrant-Whyte2006RAM,Bailey2006RAM,Cadena2016TRO,Bresson2017TIV}.
VINS can be considered as an instance of SLAM (using particular visual and inertial sensors) and 
broadly include the visual-inertial (VI)-SLAM~\cite{Lupton2012TRO,Leutenegger2014IJRR,Shen2015ICRA} and the visual-inertial odometry (VIO)~\cite{Mourikis2007ICRA,Hesch2013TRO,Li2013IJRR,Li2014,Huang2014ICRA,Huai2018IROS}. 
The former jointly estimates the feature positions and the camera/IMU pose that together form the state vector, 
whereas the latter does not include the features in the state but still utilizes the visual measurements to impose motion constraints between the camera/IMU poses. 
In general, by performing mapping (and thus loop closure), 
the VI-SLAM gains the better accuracy from the feature map and the possible loop closures while incurring higher computational complexity than the VIO, 
although different methods have been proposed to address this  issue~\cite{Shen2015ICRA,Leutenegger2014IJRR,Usenko2016ICRA,Mourikis2007ICRA,Mourikis2009TRO,Li2013RSS}.
However, VIO estimators are essentially odometry (dead reckoning) methods whose localization errors may grow unbounded
unless some global information (e.g., GPS or {\em a priori} map)  or constraints to previous locations (i.e., loop-closures) are used.
Many approaches leverage feature observations  from different keyframes to limit drift over the trajectory \cite{Nerurkar2014ICRA,Leutenegger2014IJRR}.
Most have a two-thread system that optimizes a small window of ``local'' keyframes and features limiting drift in the short-term, 
while a background process optimizes a long-term sparse pose graph containing loop-closure constraints enforcing long-term consistency~\cite{Mur-Artal2017RAL,Liu2018CVPR,Qin2018TRO,Qin2018RELOC}.
For example, VINS-Mono~\cite{Qin2018TRO,Qin2018RELOC} uses loop-closure constraints 
in both the local sliding window and in the global batch optimization.
Specifically, during the local optimization, feature observations from keyframes provide implicit loop-closure constraints, 
while the problem size remains small by assuming the keyframe poses are perfect (thus removing them from optimization).

In particular, whether or not performing loop closures in VINS either via mapping~\cite{Mur-Artal2017RAL,Schneider2018RAL,DuToit2017ICRA} 
and/or place recognition~\cite{Galvez2012TRO,Lowry2016TRO,Latif2014RSS,Latif2017RAS,Han2018AURO,Merrill2018RSS} is one of the key differences between VIO and SLAM.
While it is essential to utilize loop-closure information to enable bounded-error VINS performance,
it  is challenging due to the inability to remain computationally efficient 
without making inconsistent assumptions such as treating keyframe poses to be true, or reusing information.
To this end, 
a hybrid estimator was proposed in~\cite{Mourikis2008CVPRW}  that used the MSCKF to perform real-time local estimation, 
and triggered global BA on loop-closure detection.
This allows for the relinearization and inclusion of loop-closure constraints in a consistent manner, while requiring substantial additional overhead time where the filter waits for the BA  to finish.
% Using the information matrix calculated during batch optimization, they replaced both the mean and covariance of the MSCKF after optimization ensuring consistency.
%
Recently, \citet{Lynen2015RSS} developed a large-scale map-based VINS that uses a compressed prior map containing feature positions and their uncertainties and employs the matches to features in the prior map to constrain the estimates globally.
\citet{DuToit2017ICRA} exploited the idea of Schmidt KF~\cite{Schmidt1966ACS} and developed a Cholesky-Schmidt EKF, 
which, however, uses {\em a prior} map with its full uncertainty 
and relaxes all the correlations between the mapped features and the current state variables;
while our latest Schmidt-MSCKF~\cite{Geneva2019ICRA} integrates loop closures in a single thread.
Moreover, the recent point-line VIO~\cite{Zheng2018ICRA} treats the 3D positions of marginalized keypoints as ``true'' for  loop closure, which may  lead to inconsistency.
%

%%%%%%%%%%%%%%%%%%%%%%%%%%%%%%%%%%%%%%%%%%%%%%%%%%%%%%%%%%%%%%%%%%%%
%%%%%%%%%%%%%%%%%%%%%%%%%%%%%%%%%%%%%%%%%%%%%%%%%%%%%%%%%%%%%%%%%%%%
%%%%%%%%%%%%%%%%%%%%%%%%%%%%%%%%%%%%%%%%%%%%%%%%%%%%%%%%%%%%%%%%%%%%
\subsection{Direct vs. Indirect Visual Processing}

Visual processing pipeline is one of the key  components to any VINS,
responsible for transforming dense imagery data to motion constraints  that can be incorporated into the estimation problem,
whose algorithms can be categorized as either direct or indirect upon the visual residual models used.
Seen as the classical technique, indirect methods~\cite{Mourikis2007ICRA,Li2013IJRR,Hesch2014IJRR,Leutenegger2014IJRR,Eckenhoff2019IJRR}  extract and track point features in the environment, while using geometric reprojection constraints during estimation.
An example of a current state-of-the-art indirect visual SLAM is the ORB-SLAM2~\cite{Mur-Artal2015TRO,Mur-Artal2017RAL}, 
which performs graph-based optimization of camera poses using information from 3D feature point correspondences.

In contrast, direct methods~\cite{Forster2014ICRA,Engel2014ECCV,Usenko2016ICRA,Eckenhoff2017ICRA} utilize raw pixel intensities in their formulation and allow for inclusion of a larger percentage of the available image information.
LSD-SLAM~\cite{Engel2014ECCV} is an example of a state-of-the-art direct visual-SLAM 
which optimizes the transformation between pairs of camera keyframes based on minimizing their intensity error.
Note that this approach also optimizes a separate graph containing keyframe constraints to allow for the incorporation of highly informative loop-closures to correct drift over long trajectories.
This work was later extended from a monocular sensor to stereo and omnidirectional cameras for improved accuracy~\cite{Engel2015IROS,Caruso2015IROS}.
Other popular direct methods include \cite{Engel2018TPAMI} and \cite{Wang2017ICCV} which estimate keyframe depths along with the camera poses in a tightly-coupled manner, offering low-drift results.
Application of direct methods to VINS has seen recent attention due to their ability to robustly track dynamic motion even in low-texture environments.
For example,
\citet{Bloesch2015IROS,Bloesch2017IJRR} used a patch-based direct method to provide updates with an iterated EKF;
\citet{Usenko2016ICRA} introduced a sliding-window VINS based on the discrete preintegration and direct image alignment;
\citet{Ling2016ICRA,Eckenhoff2017ICRA} integrated direct image alignment with different IMU preintegration~\cite{Forster2015RSS,Shen2015ICRA,Eckenhoff2016WAFR} for  dynamic motion estimation.

While direct image alignments require a good initial guess and high frame rate due to the photometric consistency assumption,
indirect visual tracking consumes extra computational resources on extracting and matching features. 
Nevertheless,  indirect methods are more widely used in practical applications due to its maturity and robustness,
but direct approaches have potentials in textureless scenarios. 
%as they are operated directly on the pixel level.

\subsection{Inertial Preintegration}

\citet{Lupton2012TRO} first developed the IMU preintegration, a computationally efficient alternative to the standard inertial measurement integration,
which peforms the discrete integration of the inertial measurement dynamics in a \textit{local} frame of reference, 
thus preventing the need to reintegrate the state dynamics at each optimization step.
While this addresses the computational complexity issue, this method suffers from singularities due to the use of Euler angles in the orientation representation.
To improve the stability of this preintegration, an on-manifold representation was introduced in~\cite{Forster2015RSS,Forster2017TRO} 
which presents a singularity-free orientation representation on the $SO(3)$ manifold, incorporating the IMU preintegration into graph-based VINS.

While \citet{Shen2015ICRA} introduced preintegration in the continuous form, they still discretely sampled the measurement dynamics without offering closed-form solutions, which  left a significant gap in the theoretical completeness of preintegration theory from a continuous-time perspective.
%Albeit, \citet{Qin2018TRO} later extended to a robust tightly-coupled monocular visual-inertial localization system.
As compared to the discrete approximation of the preintegrated measurement and covariance calculations used in previous methods, 
in our prior work \cite{Eckenhoff2016WAFR,Huai2019IJRR}, 
we have derived the closed-form solutions to both the  measurement and covariance preintegration equations
and showed that these solutions offer improved accuracy over the discrete methods, especially in the case of highly dynamic motion.

\subsection{State Initialization}

Robust, fast initialization to provide of accurate initial state estimates is crucial to bootstrap real-time VINS estimators, 
which is often solved in a linear closed form~\cite{Kneip2011IROS,DongSi2012IROS,Martinelli2012TRO,Martinelli2014IJCV,Lippiello2013MED,Yang2017TASE,Shen2014ISER}.
In particular, 
\citet{Martinelli2014IJCV} introduced a closed-form solution to the monocular visual-inertial initialization problem
and later extended to the case where gyroscope bias calibration is also included~\cite{Kaiser2017RAL} as well as to the cooperative scenario~\cite{Martinelli2018CoRR}. 
These approaches fail to model the uncertainty in inertial integration since they rely on the double integration of IMU measurements over an extended period of time. 
\citet{Faessler2015ICRA} developed a re-initialization and failure recovery algorithm based on SVO~\cite{Forster2014ICRA} within a loosely-coupled estimation framework, while an additional downward-facing distance sensor is required to recover the metric scale. 
\citet{Mur-Artal2017RAL} introduced a high-latency  (about 10 seconds) initializer built upon their ORB-SLAM~\cite{Mur-Artal2015TRO},
which computes initial scale, gravity direction, velocity and IMU biases with the visual-inertial full BA given a set of keyframes from ORB-SLAM. 
In~\cite{Yang2017TASE,Shen2014ISER} a linear method was recently proposed for noise-free cases,
by  leveraging relative rotations obtained by short-term IMU (gyro) pre-integration but without modeling the gyroscope bias,
which may be unreliable in real world in particular when distant visual features are observed. 
%

%!TEX root = main.tex
%-----------------------------------------------------------------------------------------------------
\section{Sensor Calibration} \label{sec:calibration}

When fusing the measurements from different sensors, it is critical to determine in high precision both the {\em spatial} and {\em temporal} sensor calibration parameters. 
In particular, we should know accurately the \textit{rigid-body transformation} between the camera and the IMU in order to correctly fuse motion information extracted from their measurements. 
In addition, due to improper hardware triggering, transmission delays, and clock synchronization errors, the timestamped sensing data of each sensor may disagree and thus, a timeline misalignment (time offset) between visual and inertial measurements might occur, which will eventually lead to unstable or inaccurate state estimates. It is therefore important that these \textit{time offsets} should also be calibrated. 
The problem of sensor calibration of the spatial and/or temporal parameters  has been the subject of many recent VINS research efforts~\cite{Mirzaei2008TRO,Jones2011IJRR,Kelly2011IJRR,Furgale2013IROS,Li2014IJRRa,Qin2018arXiv}. 
For example,
\citet{Mirzaei2008TRO} developed an EKF-based spatial calibration between the camera and IMU. 
Nonlinear observability analysis~\cite{Hermann1977TAC} for the calibration parameters was performed to show that 
these parameters are observable given random motion. 
% %
Similarly, \citet{Jones2011IJRR} examined the identifiability of the spatial calibration of the camera and IMU 
based on indistinguishable trajectory analysis and developed a filter based online calibration  on an embedded platform. 
% %
\citet{Kelly2011IJRR} solved for the rigid-body transformation between the camera and IMU  
by aligning rotation curves of these two sensors via an ICP-like matching method.

Many of these research efforts have been focused on \textit{offline} processes 
that often require additional calibration aids (fiducial tags)~\cite{Furgale2013IROS,Rehder2016ICRA,Fleps2011IROS,Nikolic2016ISJ,Mirzaei2008TRO}.
In particular, as one of the state-of-the-art approaches, the {\em Kalibr} calibration toolbox~\cite{Furgale2013IROS,Rehder2016ICRA} 
uses a continuous-time basis function representation~\cite{Furgale2015IJRR} of the sensor trajectory 
to calibrate both the extrinsics and intrinsics of a multi-sensor system in  a batch fashion.
As this B-spline representation allows for the direct computation of expected local angular velocity and local linear acceleration,
the difference between the expected and measured inertial readings serve as errors in the batch optimization formulation.
A downside of offline calibration is that it must be performed every time a sensor suite is reconfigured.
For instance, if a sensor is removed for maintenance and returned, errors in the placement could cause poor performance,  
requiring a time-intensive recalibration.

Online calibration methods, by contrast, estimate the calibration parameters during every operation of the sensor suite, 
thereby making them more robust to and easier to use in such scenarios. 
\citet{Kim2018TASE} reformulated the IMU preintegration~\cite{Lupton2012TRO,Forster2015RSS,Eckenhoff2016WAFR} by transforming the inertial readings from the IMU frame into a second frame.
This allows for calibration between IMUs and other sensors (including other IMUs), 
but does not include temporal calibration and also relies on computing angular accelerations from gyroscope measurements.   
\citet{Li2014IJRRa} performed navigation with simultaneous calibration of both the spatial and temporal extrinsics between a single IMU-camera pair in a filtering framework for use on mobile devices,
which  was later extended to include the intrinsics of both the camera and the IMU~\cite{Li2014ICRA}. 
%while considering a single IMU case.
%
\citet{Qin2018arXiv} extended their prior work on batch-based monocular VINS~\cite{Qin2018TRO} to include the time offset between the camera and IMU by interpolating the locations of features on the image plane.
\citet{Schneider2019ISJ} proposed the observability-aware online calibration utilizing the most informative motions.
While we recently have also analyzed the degenerate motions of  spatiotemporal  calibration~\cite{Yang2019RAL},
it is not fully understood how to optimally model intrinsics and simultaneously calibrate them along with extrinsics~\cite{Rehder2017ISJ,Nikolic2016}. 
%

%-----------------------------------------------------------------------------------------------------
\section{Observability Analysis} \label{sec:obs}

%%%%%%%%%%%%%%%%%%%
System observability plays an important role in the design of consistent state estimation~\cite{Huang2012thesis},
which examines whether the information provided by the available measurements is sufficient for estimating the state/parameters without ambiguity~\cite{Brogan1991,Bar-Shalom2001,Hermann1977TAC}. 
When a system is observable, the observability matrix is invertible, 
which is also closely related to the Fisher information  (or covariance) matrix~\cite{Huang2011IROS,Huang2017SCL}.
Since this matrix describes the information available in the measurements, 
by studying its nullspace we can gain insights about the directions in the state space along which the estimator should acquire information. 
In our prior work~\cite{Huang2008ICRA,Huang2008ISER,Huang2010IJRR,Huang2009ICRA,Huang2013TRO,Huang2011AURO,Huang2011IROS,Huang2014ICRA}, 
we have been the first to design observability-constrained  (OC) consistent estimators for robot localization problems. 
Since then, significant research efforts have been devoted to the observability analysis of VINS (e.g., \cite{Hesch2013WAFR,Li2012ICRA,Yang2018ICRA,Yang2019TRO}). 

In particular, VINS nonlinear observability analysis has been studied using different nonlinear system analysis techniques.
For example, 
\citet{Jones2011IJRR,Hernandez2015ICRA}  the system's indistinguishable trajectories~\cite{Isidori1995}  were examined from the observability perspective. 
By employing the concept of continuous symmetries as in~\cite{Martinelli2011TRO}, 
\citet{Martinelli2012TRO} analytically derived the closed-form solution of VINS 
and identified that IMU biases, 3D velocity, global roll and pitch angles  are observable. 
He has also examined the effects of degenerate motion~\cite{Martinelli2013IROS}, minimum available sensors~\cite{Martinelli2014ICRA}, cooperative VIO~\cite{Martinelli2018CoRR} and unknown inputs~\cite{Martinelli2017BOOK,Martinelli2018TAC} on the system observability. 
Based on the Lie derivatives and observability matrix rank test~\cite{Hermann1977TAC}, 
\citet{Hesch2014IJRR} analytically showed that the monocular VINS has 4 unobservable directions, corresponding to  the global yaw and the global position of the exteroceptive sensor. 
\citet{Guo2013iros} extended this method to the RGBD-camera aided INS that preserves the same unobservable directions if both point and plane measurements are available.
With the similar idea, 
in~\cite{Mirzaei2007IROS,Kelly2011IJRR,Guo2013ICRA}, the observability of IMU-camera (monocular, RGBD) calibration was analytically studied, which shows  that the extrinsic transformation between the IMU and  camera is observable given generic motions.
Additionally, 
in~\cite{Panahandeh2013IROS,Panahandeh2016JIRS}, the system with a downward-looking camera measuring point features from horizontal planes was shown to have the observable global $z$ position of the sensor.

%%%%
As in practice VINS estimators are typically built upon the linearized system, 
what is practically more  important is to perform  observability analysis for the linearized VINS.
%whose observability properties can be exploited when designing a state estimator.
%
%%
In particular, the observability matrix~\cite{Maybeck1979,Chen1990IECON} for the linearized VINS system 
over the time interval $[k_o ~ k]$ 
has the nullspace (i.e., unobservable subspace) that {\em ideally} spans {\em four} directions:
\begin{align}% \label{eq:obs-matrix}
\scalemath{.95}{
\mathbf M = \begin{bmatrix}
\mathbf H_{k_o} \\
\mathbf H_{k_o+1} \bm\Phi_{k_o} \\
\vdots\\
\mathbf H_{k} \bm\Phi_{k-1}\cdots \bm\Phi_{k_o}
\end{bmatrix}
\overset{\mathbf{MN=0}}{\Longrightarrow}
\mathbf N = \begin{bmatrix}
\mathbf 0_3 & \mathbf C(^I_G\bar{\mathbf q}_k) {^G\mathbf g} \\
\mathbf 0_3 & \mathbf 0_3 \\
\mathbf 0_3 & -\lfloor ^G\mathbf v_k \times \rfloor {^G\mathbf g} \\
\mathbf 0_3 & \mathbf 0_3 \\
\mathbf I_3 & -\lfloor ^G\mathbf p_k \times \rfloor {^G\mathbf g} \\
\mathbf I_3 & -\lfloor ^G\mathbf p_{f} \times \rfloor {^G\mathbf g} 
\end{bmatrix} 
} \label{eq:nullspace}
\end{align}
Note that the first block column of $\mathbf N$ in~\eqref{eq:nullspace} 
corresponding to the global translation
while the second block column corresponds to the 
and global rotation about the gravity vector~\cite{Li2012ICRA,Li2013IJRR,Hesch2013WAFR,Hesch2014IJRR}.
When designing a nonlinear estimator for VINS, we would like the system model employed by the estimator to have
an unobservable subspace spanned by these directions.
However, this is not the case for the standard EKF as shown in~\cite{Li2012ICRA,Li2013IJRR,Hesch2013WAFR,Hesch2013TRO,Hesch2014IJRR}. 
In particular, the standard EKF linearized system, 
which linearizes system and measurement functions at the {current} state estimate,
has an unobservable subspace of {\em three}, instead of four dimensions.
This implies that the filter gains non-existent information from available measurements,  leading to inconsistency.
To address this issue,
%performed observability analysis for the linearized VINS (without considering biases) and 
the first-estimates Jacobian (FEJ) idea~\cite{Huang2008ISER} was adopted to improve MSCKF consistency~\cite{Li2012ICRA,Li2013IJRR},
and the OC methodology~\cite{Huang2010IJRR} was employed in developing the OC-VINS~\cite{Hesch2013WAFR,Hesch2013TRO,Kottas2012ISER}.
%conducted observability analysis for the linearized VINS with full states (including IMU biases) and analytically showed the system unobservable directions by finding the right null space of the observability matrix~\cite{Hesch2013TRO}. 
%Based on this analysis, 
We recently have also developed the robocentric VIO (R-VIO)~\cite{Huai2018IROS,Huai2019IJRR} 
which preserves proper observability properties independent of linearization points.

%-----------------------------------------------------------------------------------------------------%-----------------------------------------------------------------------------------------------------
\section{Discussions and Conclusions} \label{sec:concl}

As inertial and visual sensors are becoming ubiquitous,  
visual-inertial navigation systems (VINS) have incurred significant research efforts
and witnessed great progresses in the past decade,
fostering an increasing number of innovative applications in practice.
As a special instance of the well-known SLAM problem, 
VINS researchers have been quickly building up a rich body of literature on top of SLAM~\cite{Cadena2016TRO}.
Given the growing number of papers published in this field, 
it has become harder (especially for practitioners) to keep up with the state of the art.
Moreover, because of the particular sensor characteristics, 
it is not trivial  to  develop VINS algorithms from scratch without understanding the pros and cons of existing approaches in the literature
(by noting that each method has its own particular focus and does not necessarily explain all the aspects of VINS estimation).
All these have motivated us to provide this review on VINS, 
which, to the best of our knowledge, is unfortunately lacked in the literature
and thus should be a useful reference for researchers/engineers who are working on this problem.
Upon our significant prior work in this domain, we have strived to make this review concise but complete,
by focusing on the key aspects about building a VINS algorithm including state estimation,
%(which is at the core of any VINS algorithm), 
sensor calibration and observability analysis.

While there are significant progresses on VINS made  in the past decade, 
many challenges remain to cope with, and in the following we just list a few  open to discuss:
\begin{itemize}

\item {\em Persistent localization:}
While current VINS are able to provide accurate 3D motion tracking, but, in small-scale friendly environments,
they are not robust  enough for long-term, large-scale,  safety-critical deployments, e.g.,  autonomous driving, 
in part due to resource constraints~\cite{Li2013RSS,Zhang2017RSS,Li2014}. 
As such, it is demanding to enable persistent VINS even in challenging conditions (such as bad lighting and motions), 
e.g., by efficiently integrating loop closures or building and utilizing novel maps.

\item {\em Semantic localization and mapping:}
Although geometric features such as points, lines and planes~\cite{Yang2018ICRA,Yang2019ICRAa} are primarily used in current VINS for localization, 
these handcrafted features may not be work best for navigation,
and it is of importance to be able to learn best features for VINS by leveraging recent advances of deep learning~\cite{dl-book}.
Moreover, a few recent research efforts have attempted to endow VINS with semantic understanding of  environments~\cite{Dong2017CVPR,Bowman2017ICRA,Fei2018arXiv,Lianos2018ECCV},
which  is only sparsely explored  but holds great potentials.

\item {\em High-dimensional object tracking:}
When navigating in dynamic complex environments, 
besides high-precision localization, it is often necessary to detect, represent, and track moving objects 
that co-exist in the same space in real time, for example, 3D object tracking in autonomous navigation~\cite{Li2018ECCV,Eckenhoff2018ICRAws,Eckenhoff2019RAL}.

\item {\em Distributed cooperative VINS:}
Although cooperative VINS have been preliminarily studied in~\cite{Melnyk2012ICRA,Martinelli2018CoRR}, 
it is still challenging to develop real-time distributed VINS, e.g., for crowd sourcing operations. 
Recent work on cooperative mapping~\cite{Guo2016ICRA,Guo2018TRO} may shed some light on how to tackle this problem.

\item {\em Extensions to different aiding sensors:}
While optical cameras are seen an ideal aiding source for INS in many applications,
other aiding sensors may more proper for some environments and motions,
for example, acoustic sonars may be instead used in underwater~\cite{Yang2017ICRA};
low-cost light-weight LiDARs may work better in environments, e.g., with poor lighting conditions~\cite{Hesch2010ICRA,Geneva2018IROS};
and event cameras~\cite{Lichtsteiner2008JSSC,Liu2010CON} may better capture dynamic motions~\cite{Zhu2017CVPR,Mueggler2018TRO}.
Along this direction, we should investigate in-depth VINS extensions of using different aiding sources for applications at hand.

\end{itemize}

%%%%---------------------------------------
\bibliographystyle{IEEEtranN}  
\bibliography{xins,calibration,rpng}

\end{document}